%% file: main.tex
\documentclass[letterpaper, 10 pt, conference]{ieeeconf}  
\IEEEoverridecommandlockouts                             
\usepackage{graphics} % for pdf, bitmapped graphics files
\usepackage{epsfig} % for postscript graphics files
\usepackage{times} % assumes new font selection scheme installed
\usepackage{amsmath} % assumes amsmath package installed
\usepackage{amssymb}  % assumes amsmath package installed
\usepackage{xcolor}
\usepackage{bm}
\usepackage{cancel}
\usepackage{color}
\usepackage{setspace}
\usepackage{wrapfig}
\usepackage{tikz}
\usetikzlibrary{shapes,arrows.meta,calc}
\usetikzlibrary{positioning}
\usepackage{caption}
\usepackage{subcaption}
\usepackage{booktabs}
\usepackage{multirow}
\usepackage{multicol}
\usepackage{soul}
\usepackage{cite}
\usepackage[normalem]{ulem}
\usepackage{xcolor}
\usepackage{mathtools}
\usepackage{stfloats}

\let\labelindent\relax % Use for IEEE templates to suppress the already defined \labelindent
\usepackage{enumitem}
\usepackage{tabularx}
\usepackage{courier}
\usepackage{algorithm}
\usepackage{algorithmic}
\usepackage{makecell}
\usepackage{url}
\usepackage{hyperref}
\usepackage{units}

\usepackage{pifont}% http://ctan.org/pkg/pifont
\usepackage{url}

\newcommand{\todocite}[1]{\textcolor{red}{[TODO(cite)]}}

\allowdisplaybreaks

\newcommand{\norm}[1]{\lVert #1 \rVert}

\title{\LARGE 
\textbf{
Autotuning Bipedal Locomotion MPC with GRFM-Net for Efficient Sim-to-Real Transfer 
}
}

\author{Qianzhong Chen$^{*1}$, Junheng Li$^{*2}$, Sheng Cheng$^{3}$, Naira Hovakimyan$^{3}$, and Quan Nguyen$^{2}$
\thanks{$^*$ Equal contributions.}
\thanks{This work is supported by NASA cooperative agreement (80NSSC22M0070), NSF-AoF Robust Intelligence (2133656), NSF SLES (2331878), and USC Departmental Startup Fund.
}
\thanks{$^{1}$ Author is with the Department of Mechanical Engineering, Stanford University, USA. Email: {\tt\small qchen23@stanford.edu}
}
\thanks{$^{2}$ Authors are with the Department of Aerospace and Mechanical Engineering, University of Southern California, USA. Email: {\tt\small \{junhengl, quann\}@usc.edu}
}
\thanks{$^{3}$ Authors are with the Department of Mechanical Science and Engineering, University of Illinois Urbana-Champaign, USA. Email: {\tt\small \{chengs, nhovakim\}@illinois.edu}
}
}

\begin{document}
\maketitle
% for editing
% \thispagestyle{plain}
% \pagestyle{plain}
% %% for submission
\thispagestyle{empty}
\pagestyle{empty}

\begin{abstract}
    Bipedal locomotion control is essential for humanoid robots to navigate complex, human-centric environments. While optimization-based control designs are popular for integrating sophisticated models of humanoid robots, they often require labor-intensive manual tuning. In this work, we address the challenges of parameter selection in bipedal locomotion control using DiffTune, a model-based autotuning method that leverages differential programming for efficient parameter learning. A major difficulty lies in balancing model fidelity with differentiability. We address this difficulty using a low-fidelity model for differentiability, enhanced by a Ground Reaction Force-and-Moment Network (GRFM-Net) to capture discrepancies between MPC commands and actual control effects. We validate the parameters learned by DiffTune with GRFM-Net in hardware experiments, which demonstrates the parameters' optimality in a multi-objective setting compared with baseline parameters, reducing the total loss by up to 40.5$\%$ compared with the expert-tuned parameters. The results confirm the GRFM-Net's effectiveness in mitigating the sim-to-real gap, improving the transferability of simulation-learned parameters to real hardware.
\end{abstract}

% \begin{center}
%     SUPPLEMENTARY MATERIAL
% \end{center}
% Video: https://youtu.be/25Z7iAkZ5xw \\
% Code: https://github.com/HovakimyanResearch/L1-Mambo

\input{introduction}

\input{literature}
\input{background}

\input{approach}

\input{results}

\input{conclusion}

% \newpage 
\bibliographystyle{IEEEtran}
\bibliography{ref}

\end{document}

%% file: introduction.tex
\section{Introduction}

%% motivation of studying bipedal robot locomotion
Bipedal locomotion control involves studying and developing algorithms that enable bipedal/humanoid robots to walk, balance, and navigate various human-centric terrains. This field is motivated by the desire to create robots that can operate and perform tasks that are challenging or dangerous for humans, and assist in areas such as industrial plants, rehabilitation, and disaster response missions.

In the past decade, there has been growing interest in studying bipedal locomotion, including footstep planning \cite{deits2014footstep,xiong2021global}, autonomous navigation \cite{li2023autonomous}, locomotion control \cite{kuindersma2016optimization,nguyen2018dynamic,kim2020dynamic}, and safety guarantees \cite{ahmadi2021risk,shamsah2023integrated}. Among these studies, bipedal locomotion control remains the most fundamental topic of humanoid robotics research.
% Recent studies have shown promising results on robust locomotion via Model-predictive Control \cite{gu2024walking,khazoom2024tailoring,daneshmand2021variable}. 
% The numerical-optimization-based control strategy remains a pertinent and favorable approach. 
Gradient-based optimizations have always been popular approaches for bipedal locomotion planning and control, including methods such as Whole-Body Control (WBC) \cite{kim2020dynamic,klemm2020lqr}, Model Predictive Control (MPC) \cite{daneshmand2021variable,dantec2022whole,gu2024walking}, and direct Trajectory Optimization (TO) \cite{posa2013direct,apgar2018fast}.
As an early bipedal locomotion technique, the Zero-Moment-Point (ZMP) method combines with the Linear Inverted Pendulum Model (LIPM) for footstep planning with preview control \cite{kajita2003biped}. Hybrid Zero Dynamics (HZD) later emerged as a favorable approach, integrating the robot's dynamics with impact modeling to achieve highly dynamic bipedal locomotion \cite{westervelt2003hybrid,sreenath2011compliant}. More recently, force-based modeling has been employed in bipedal locomotion MPC and TO, utilizing models such as centroidal dynamics \cite{orin2013centroidal, kuindersma2016optimization} and single rigid-body dynamics \cite{li2021force}. These approaches enable the generation of compliant motions, enhancing the robot's ability to adapt to external forces and terrain perturbations.
%% challenges with optimization-based control in bipedal locomotion :
With advancements in efficient solvers and computational hardware, these optimization-based control designs have become increasingly sophisticated, featuring greater model complexity, multi-objective optimization, and higher-dimensional optimization spaces. Employing full-order dynamics modeling in finite-horizon optimization problems becomes more and more common \cite{dantec2022whole, khazoom2024tailoring}.
As a result, these complex controllers often require extensive, unintuitive, and even expert-level parameter tuning to achieve reasonable performance \cite{wensing2023optimization}, which raises the barrier for newcomers and complicates the efficient development of new control algorithms.

%%%%% dynamics illustration %%%%%
\begin{figure}[!t]
\vspace{0.2cm}
		\center
		\includegraphics[clip, trim=0.5cm 1cm 0.5cm 1cm, width=1\columnwidth]{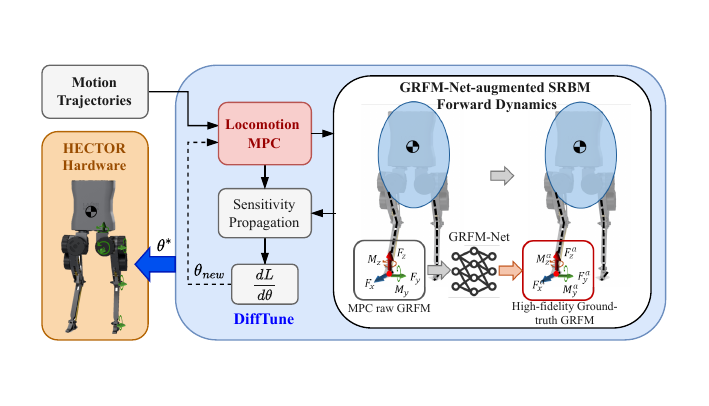}
		\caption{ Autotuning Bipedal MPC Parameters Through DiffTune and GRFM-Net. Supplementary video: \url{https://youtu.be/bfrBW2hIT1M} }
		\label{fig:Dynamics}
		\vspace{-0.4cm}
\end{figure}

%%%%%%%%%%%%%%%%%%%%%%%%%%%%%%%%%%%%%%%%%%%%%%%%%

To address the challenges in the parameter selection, previous works have explored Bayesian optimization (BO) to auto-tuning controller parameters used for bipedal locomotion and control~\cite{calandra2014experimental,rai2018bayesian,yeganegi2021robust,yang2022bayesian}. {Bayesian optimization is a model-free optimization algorithm and is suitable for finding global optima for black-box optimization functions. However, BO scales poorly with the dimensionality of the parameter space, where the search space grows exponentially.} A recent work~\cite{yuan2019bayesian} addresses the scalability issue by applying domain knowledge for partitioning the parameter space and using alternating BO algorithms to iteratively update the parameters on the subspace. 
In contrast to the model-free BO approach for parameter learning, model-based autotuning methods~\cite{parwana2021recursive,cheng2022difftune,cheng2023difftuneplus,tao2023DT-MPC,kumar2021diffloop} offer more efficient parameter search. Using autodifferentiation, these methods can easily acquire the first-order information (compared with BO which only uses the zeroth-order information) that can efficiently guide the parameter update. Notably, the state-of-the-art model-based auto-tuning method, DiffTune~\cite{cheng2022difftune}, has achieved better tracking accuracy than BO in fewer trials in auto-tuning a nonlinear quadrotor control system.

%% our solution/paper highlight
In this work, we apply DiffTune to learn the optimal parameters of a linear locomotion MPC for bipedal walking with different trajectories and to demonstrate the feasibility of autotuning high-dimensional optimization problems for under-actuated legged locomotion. 
A dilemma in applying DiffTune (or other differentiable-programming-based learning methods) is the tradeoff between fidelity and differentiability of the dynamics model for legged robots. 
One option is the highly accurate full-order dynamics, which, though, is not easily differentiable due to its complexity, including discrete variables from ground contact models and the need for inverse kinematics solutions for swing leg control. Another option is the Single Rigid Body Model (SRBM), which offers differentiability but lacks fidelity due to simplifying assumptions. 
To tackle this dilemma, we take the SRBM for its differentiability and seek to improve its fidelity with a Ground Reaction Force-and-Moment Network (GRFM-Net). The GRFM-Net is designed to capture the discrepancies between the MPC command and the actual control effect in a data-driven manner, accounting for actuator dynamics, contact interaction, and leg transmission mechanics. It improves the simulation accuracy while maintaining the differentiability for parameter learning. 
We train the GRFM-Net with data collected from a high-fidelity (nondifferentiable) simulator due to the hardware and sensor limits in obtaining high-quality data. With the fidelity enhanced by GRFM-Net, DiffTune can find optimal parameters that improve the trajectory tracking accuracy when evaluated in hardware experiments. In contrast, DiffTune without GRFM-Net showed significantly worse performance, especially in more complex trajectories, highlighting the critical role of GRFM-Net in mitigating the sim-to-real gap and ensuring robust parameter learning across varying levels of task difficulty.
% To improve fidelity, the Augmented Actuator Network (AAN) is introduced, using a data-driven approach to better capture the discrepancies between commanded and actual control actions, accounting for actuator and contact dynamics. This improves simulation accuracy while maintaining differentiability for learning.

% \textcolor{blue}{Qianzhong: Our solution with DiffTune and usage of actuator net for augmenting the fidelity of SRB dynamics.} \\ 
% To mitigate the challenge of bipedal robot controller parameters selection, we proposed to use DiffTune (cite), a widely used auto-tuning framework that iteratively lowering the control loss by conducting sensitivity analysis w.r.t. controller parameters. DiffTune requires a differentiable simulation environment and the following simple SRB dynamic model, which leads to critical sim-to-real gap. To address it, we trained an augmented actuator net (AAN) to map the MPC controller commands into real ground reactions, elevating the fidelity of simulator largely in a data-driven way.  

%%%% main contributions %%%%
Our contributions are summarized as follows: (i) We apply DiffTune to learn model-based locomotion MPC's parameters for bipedal walking, which leverages the first-order information to achieve more efficient parameter learning than commonly used Bayesian optimization that only uses zeroth-order information. (ii) We propose GRFM-Net, a data-driven method to enhance the fidelity of a differentiable low-fidelity simulator, which reduces the sim-to-real gap when deploying the simulation-learned parameters to a real physical system.

%% file: background.tex
\section{Background} \label{sec: tech background}

% \subsection{Hector dynamics and controller [Junheng]}
%%%%%%%%%%%%%%%%%%%%%%%%%%%%%%%%%%%%%%%%%%%%%%%%%
\subsection{HECTOR Bipedal Robot Dynamics Modelling}
\label{subsec:hector}
% \jli{I will shorten the background, maybe omit sections such as the swing leg control and equation (3-4) to save space}
We first introduce the robot and the dynamics models used in this work. HECTOR V2 bipedal robot is a successor of the HECTOR humanoid introduced in \cite{li2023dynamic}. HECTOR biped consists of 5-DoF legs with ankle pitch actuation, shown in Fig.~\ref{fig:Dynamics}. Standing at 70 $\unit{cm}$ and weighing 12 $\unit{kg}$, the biped has maximum knee joint torque of 67 $\unit{Nm}$.

HECTOR's full-order dynamics equation of motion in generalized coordinates is described as follows:
\begin{align}
\label{eq:fullDynamics}
    \mathbf{H}(\mathbf{q})\ddot{\mathbf{q}} + \mathbf{C}(\mathbf{q}, \dot{\mathbf{q}}) = \mathbf{\Gamma} + \bm{J}_i(\mathbf{q})^\intercal \bm{\lambda}_i ,
\end{align}
where $\mathbf{H} \in \mathbb{R}^{16 \times 16}$ is the mass-inertia matrix and $\mathbf{C} \in \mathbb{R}^{16}$ is the joint-space bias force term. 
The joint-space state {${\mathbf{q}} \in \mathbb{R}^{16}$} includes 
float-base CoM position $\bm p_c$, Euler angles $\mathbf e $, and joint positions $\bm q_j$; $\mathbf{\Gamma} = [\mathbf{0}_6; \bm \tau_j]$ represents the actuation in the generalized coordinate; $\bm \lambda_i $ and $\bm{J}_i$ represent the external force applied to the system and its corresponding Jacobian matrix; $i \in \{0,1\}$ indexes the left and right leg (foot).

% Using full-order dynamics models in finite-horizon optimal control problems (\textit{e.g.}, online MPC) often poses challenges related to computational burden and efficiency. These issues typically necessitate problem linearization within solver mechanisms, such as Sequential Quadratic Programming (SQP) and Differential Dynamics Programming (DDP). To achieve acceptable computation speeds for real-time deployment, these methods often require specific tailoring, which may involve trade-offs in solution accuracy \cite{galliker2022planar,khazoom2024tailoring}. Therefore, simplified dynamics are viable options for achieving reliable and high-frequency solutions in optimization-based control \cite{wensing2023optimization}.

When using the full-order nonlinear dynamics models in finite-horizon optimal control (e.g., online MPC), one often faces computational challenges, which require linearization within solvers like Sequential Quadratic Programming and Differential Dynamics Programming. For real-time deployment, these methods also need specific tailoring \cite{galliker2022planar,khazoom2024tailoring}. Thus, simplified dynamics still remain a practical choice for achieving reliable, high-frequency solutions in optimization-based control \cite{wensing2023optimization}.

Our prior work \cite{li2021force} proposed a force-and-moment-based single rigid-body dynamics model (SRBM) approach in bipedal locomotion MPC. The SRBM, illustrated in Figure~\ref{fig:Dynamics}, neglects the light-weight lower thigh and calf links and assumes the upper body and hips as a rigid body with a constant moment of inertia (MoI). The rigid body is actuated by world-frame ground reaction forces and moments (GRFM), denoted by $\mathbf F_i$ and $ \mathbf M_i$, respectively. 
% From the investigation and discussion in \cite{}, the body frame GRFM designed for biped with 5 degree-of-freedom (DoF) legs consists of GRF $\prescript{}{\mathcal{B}}{\mathbf F_i} = [\prescript{}{\mathcal{B}}{F_x},\: \prescript{}{\mathcal{B}}{F_y},\:\prescript{}{\mathcal{B}}{F_z}]^\intercal$ and GRM with zero x-direction moment $\prescript{}{\mathcal{B}}{\mathbf M_i} = [0,\:\prescript{}{\mathcal{B}}{M_y},\:\prescript{}{\mathcal{B}}
% {M_z}]^\intercal$. 
The equations of motion are described as
\begin{subequations}\label{eq:srbd_eom}
    \begin{equation}
        m (\ddot{\bm p}_c + \bm g) = \Sigma^1_{i = 0} \mathbf F_i ,
    \end{equation}
    \begin{equation}  
\frac{d}{dt}(\prescript{}{\mathcal{G}}{\mathbf I} \bm \omega) =
\Sigma^1_{i = 0} \{ \bm r_{i}^{f} \times \mathbf F_i + \mathbf M_i  \},
    \end{equation}
\end{subequations}
where $m$ is the mass of the robot's rigid body; $\bm g$ is the gravity vector; $\prescript{}{\mathcal{G}}{\mathbf I}$ is the world-frame MoI; $\bm \omega$ is the CoM angular velocity vector; $\bm r_{i}^{f}$ is the distance vector from CoM $\bm p_c$ to the center of foot $i$. 

% This SRBM is used as the linear dynamics in a convex MPC for bipedal locomotion control. 

\subsection{SRBM-based Bipedal Locomotion MPC}
Whenever a foot is in contact (i.e., stance foot), the SRBM-based MPC will solve for optimal GRFM for the corresponding foot to remain balanced.
In the bipedal locomotion MPC, we linearize the SRBM and design a convex MPC formulation that can be solved through quadratic programming efficiently. 
We choose to include the robot CoM position $\bm p_c$, linear velocity $\dot{\bm p}_c$, Euler angles $\mathbf e$, angular velocity $\bm \omega$, and gravity vector $\bm g$ as the optimization state variables $\bm x = [\mathbf{e}; \:\bm p_c; \:\bm \omega; \:\dot{\bm p}_c; \: \bm g]$ to linearize the SRBM dynamics in \eqref{eq:srbd_eom}
and form the discrete-time state-space equation (\ref{eq:mpcDynamics}) at time step $k$. 
% \begin{align}
% \label{eq:discreteSS}
% \bm {x}_{k+1} = \bm {\hat{A}}_k\bm x_k + \bm {\hat{B}}_k\bm u_k
% \end{align}
The control input $\bm u$ includes the 3-D GRFM of both feet and $\bm u = [\mathbf F_0;\mathbf F_1;\mathbf M_0;\mathbf M_1]$. The finite-horizon optimization problem with $N$ steps can be written as
\begin{alignat}{3}
\label{eq:MPCcost}
\underset{\bm x, \bm u}{\text{minimize}} \quad & \sum_{k = 0}^{N-1} \Big\| \bm x_k-  \bm x^{\text{ref}}_k\Big\|^2 _{\bm Q} + \Big\| \bm{u}_k  \Big\|^2 _{\bm R}\\ 
    \nonumber
    \textrm{subject to:} \quad & \quad
\end{alignat}
\vspace{-0.35cm}
\begin{subequations}\label{eq: MPC constraints}
\allowdisplaybreaks
\setlength\abovedisplayskip{-3pt}
\begin{alignat}{3}
    \label{eq:mpcDynamics}
    \textrm{Dynamics: } \quad & \bm {x}_{k+1} = \bm {\hat{A}}_k\bm x_k + \bm {\hat{B}}_k\bm u_k  \\
    \label{eq:mpcFriction}
    \textrm{Friction pyramid} \quad & -\mu  {F}_{z,i,k} \leq  F_{x,i,k} \leq \mu  {F}_{z,i,k} \\
    \nonumber
    \textrm{of foot $i$:} \quad & -\mu {F}_{z,i,k} \leq  F_{y,i,k} \leq \mu  {F}_{z,i,k} \\
    \label{eq:MPCforce}
    \textrm{Force limit:} \quad  &\: 0 \leq  F_{z,i,k} \leq {F}_{\max} \\
    \label{eq:Mx}
    \textrm{Moment X:} \quad & \prescript{}{\mathcal{B}}{M_{x,i,k}} = 0 \\
    \label{eq:MPClinefoot}
    \textrm{Line foot:} \quad & -l_h F_{z,i,k} \leq M_{y,i,k} \leq l_t F_{z,i,k}.
\end{alignat}
\end{subequations}
The objective of the problem is to drive the state $\bm x$ close to the reference $\bm x^{\text{ref}}$ and minimize the control input $\bm u$. These objectives are weighted by diagonal matrices $\bm Q\in  \mathbb{R}^{15\times15}$ and $\bm R\in \mathbb{R}^{12\times12}$; $\mu$ is the ground friction coefficient; $F_{\max}$ stands for the maximum vertical force a foot can exert on the ground. When the $i$th leg is in the swing phase, $F_{\max} = 0$. Constraints (\ref{eq:Mx}-\ref{eq:MPClinefoot}) follow contact wrench cone \cite{caron2015stability} and line-foot constraints \cite{li2023dynamic}, where $l_t$ and $l_h$ are robot toe and heel lengths, respectively.  

The force-to-torque mapping of the stance leg $i$ is
\begin{align}
\bm \tau_{\text{stance}} = \left[\begin{array}{c} 
\bm J_v\\
\bm J_\omega
\end{array} \right]^\intercal
\left[\begin{array}{c} 
\mathbf F_i \\
\mathbf M_i
\end{array} \right],
\end{align}
where $\bm J_v$ and $\bm J_\omega$ are the velocity and angular velocity parts of the world-frame contact Jacobian matrices.

The swing foot is under an additional swing leg joint-space PD controller. The desired swing foot position $\bm p_f^{\text{des}}$ is generated through heuristic policies \cite{raibert1986legged}: 
\begin{gather}
    \bm p_f^{\text{des}} = \bm p_c + \frac{\dot{\bm p}_c\Delta t}{2}, 
\end{gather}
where $\Delta t$ is the swing duration.

We use an inverse kinematics solver to convert $\bm p_f^{\text{des}}$ to joint-space commands for PD control law and torque command $\bm \tau_{\text{swing}}$. 
The resulting joint torque commands for stance leg $\bm \tau_{\text{stance}}$ and swing leg $\bm \tau_{\text{swing}}$ are then used to actuate the bipedal robot. 
% The resulting joint torque commands for stance leg $i$,  $\bm \tau_{\text{stance}}$, and swing leg $i'$,  $\bm \tau_{\text{swing}}$, are then used to actuate the bipedal robot. 

%%%%%%%%%%%%%%%%%%%%%%%%%%%%%%%%%%%%%%%%%%%%%%%%%
\subsection{Controller auto-tuning and DiffTune}

We briefly review controller auto-tuning and DiffTune for a general system in the sequel.
Consider a discrete-time dynamical system
\vspace{-1mm}
\begin{equation}\label{eq: dynamics}
    \boldsymbol{x}_{j+1} = f(\boldsymbol{x}_j,\boldsymbol{u}_j), \ \boldsymbol{x}_0 \text{ given},
    \vspace{-1mm}
\end{equation}
where $\boldsymbol{x}_j$ and $\boldsymbol{u}_j $ are the state and control, respectively, with appropriate dimensions. The control is generated by a feedback controller that tracks a reference state $\boldsymbol{x}^{\text{ref}}_{j}$ such that\vspace{-1mm}
\begin{equation}
    \boldsymbol{u}_j = h(\boldsymbol{x}_j,\boldsymbol{x}^{\text{ref}}_{j},\boldsymbol{\theta}), \label{eq: feedback controller}
    \vspace{-1mm}
\end{equation}
where $\boldsymbol{\theta} \in \Theta$ denotes the controller's parameters, and $\Theta$ represents a feasible set of parameters that can be analytically or empirically determined for a system's stability.
% We assume that the state $\boldsymbol{x}_k$ can be measured directly or, if not, an appropriate state estimator is used. Furthermore, we assume that the dynamics \eqref{eq: dynamics} and controller \eqref{eq: feedback controller} are differentiable, i.e., the Jacobians $\nabla_{\boldsymbol{x}}f$, $\nabla_{\boldsymbol{u}} f$, $\nabla_{\boldsymbol{x}}h$, and $\nabla_{\boldsymbol{\theta}} h$ exist, which widely applies to general systems.

The controller's auto-tuning (or parameter learning) task adjusts $\boldsymbol{\theta}$ to minimize an evaluation criterion, denoted by a loss function $L(\cdot)$, which is a function of the desired states, actual states, and control actions over a time interval of length $T$. An illustrative example is the quadratic loss of
the tracking error and control-effort penalty, where $L(\boldsymbol{x}_{1:T},\boldsymbol{x}^{\text{ref}}_{1:T},\boldsymbol{u}_{0:T-1};\boldsymbol{\theta}) = \sum_{j=1}^{T} \norm{\boldsymbol{x}_j -\boldsymbol{x}^{\text{ref}}_{j}}^2 + \sum_{j=0}^{T-1}\lambda \norm{\boldsymbol{u}_j}^2$ 
with $\lambda>0$ being the penalty coefficient. We will use the short-hand notation $L(\boldsymbol{\theta})$ for conciseness in the rest of the paper. 

Note that we distinguish between the time indices used in auto-tuning ($j$ and $T$ for closed-loop systems) and those in MPC ($k$ and $N$ for open-loop MPC solution). For the closed-loop system, at time $j$, the instantaneous state $\boldsymbol{x}_j$ is set to initialize $\boldsymbol{x}_{k=0}$ in \eqref{eq: MPC constraints}. After solving the MPC problem,  the initial optimal control action $\boldsymbol{u}_{k=0}^\star$ is applied as $\boldsymbol{u}_{j}$ for the closed-loop system. Additionally, the horizon $T$ of auto-tuning (for performance evaluation) is generally much longer than the MPC's planning horizon $N$.

% \begin{remark}
%     The example loss function above can take a more sophisticated norm function, such as the Mahalanobis norm, which permits the assignment of weights to the state norm subject to different physical units of the state variables.
% \end{remark}

To summarize, we formulate the controller auto-tuning as the following parameter optimization problem
\begin{equation}\label{prob: controller tuning as a parameter optimization}
    \begin{aligned} 
    & \underset{\boldsymbol{\theta} \in \Theta}{\text{minimize}} && L(\boldsymbol{\theta})\\
    & \text{subject to} && \boldsymbol{x}_{j+1} = f(\boldsymbol{x}_j,\boldsymbol{u}_j), \ \boldsymbol{x}_0 \text{ given},\\
    & && \boldsymbol{u}_j \! =\! h(\boldsymbol{x}_j,{\boldsymbol{x}}_{j}^{\text{ref}},\boldsymbol{\theta}),\!\ j \in \{0,1,\dots,T-1\}.
    \end{aligned} \nonumber
    \tag{P}
\end{equation}
If the loss function $L(\cdot)$, dynamics~\eqref{eq: dynamics}, and controller~\eqref{eq: feedback controller} are all differentiable, then we are able to apply DiffTune to learn the parameters and minimize the loss.
DiffTune uses a projected gradient descent algorithm~\cite{parikh2014proximal} to update the parameters iteratively $    \boldsymbol{\theta} \leftarrow \mathcal{P}_{\Theta} ( \boldsymbol{\theta} - \beta \nabla_{\boldsymbol{\theta}}L)$,
where $\beta$ is the learning rate and $\mathcal{P}_{\Theta}$ is the projection operator that projects the updated parameter value to the feasible set $\Theta$.
To obtain the gradient $\nabla_{\boldsymbol{\theta}} L $, we start with the chain rule: 
\begin{equation}
    \nabla_{\boldsymbol{\theta}}L =\sum_{j=1}^T \frac{\partial L}{\partial \boldsymbol{x}_j} \frac{\partial \boldsymbol{x}_j}{\partial \boldsymbol{\theta}} + \sum_{j=0}^{T-1} \frac{\partial L}{\partial \boldsymbol{u}_j} \frac{\partial \boldsymbol{u}_j}{\partial \boldsymbol{\theta}}.\label{eq: decomposition of the target derivative to accepting sensitivity propagation}
\end{equation}
The gradients $\partial L / \partial \boldsymbol{x}_j$ and $\partial L / \partial \boldsymbol{u}_j$ can be determined once $L$ is chosen, and we use sensitivity propagation to obtain the sensitivity states $\partial \boldsymbol{x}_j / \partial \boldsymbol{\theta}$ and $\partial \boldsymbol{u}_j / \partial \boldsymbol{\theta}$ by:
\begin{subequations}\label{eq: sensitivity propagation}
\begin{align}
    \frac{\partial \boldsymbol{x}_{j+1}}{\partial \boldsymbol{\theta}} = & (\nabla_{\boldsymbol{x}_j} f + \nabla_{\boldsymbol{u}_j} f \nabla_{\boldsymbol{x}_j} h) \frac{\partial \boldsymbol{x}_j}{\partial \boldsymbol{\theta}} + \nabla_{\boldsymbol{u}_j} f \nabla_{\boldsymbol{\theta}} h, \label{eq: iterative Jacobian of state wrt parameter} \\
    \frac{\partial \boldsymbol{u}_j}{\partial \boldsymbol{\theta}} = & \nabla_{\boldsymbol{x}_j} h \frac{\partial \boldsymbol{x}_j}{\partial \boldsymbol{\theta}} + \nabla_{\boldsymbol{\theta}} h, \label{eq: iterative Jacobian of control wrt parameter}
\end{align}
\end{subequations}
with $\partial \boldsymbol{x}_0/\partial \boldsymbol{\theta} = 0$.
The notations $\nabla_{\boldsymbol{x}_j} f$, $\nabla_{\boldsymbol{u}_j} f$, $\nabla_{\boldsymbol{x}_j} h$, and $\nabla_{\boldsymbol{\theta}} h$ refer to the Jacobians $\partial f / \partial \boldsymbol{x}$, $\partial f / \partial \boldsymbol{u}$, $\partial h / \partial \boldsymbol{x}$, and $\partial h / \partial \boldsymbol{\theta}$ is evaluated at the state $\boldsymbol{x}_j$ and control $\boldsymbol{u}_j$. 
Since the dynamics are known, one can access $\partial f / \partial \boldsymbol{x}$ and $\partial f / \partial \boldsymbol{u}$ using autodifferentiation tools like CasADi or PyTorch. The control relevant Jacobians $\partial h / \partial \boldsymbol{x}$ and $\partial h / \partial \boldsymbol{\theta}$ are from differentiating the MPC control policy, which is detailed in~\cite{tao2023DT-MPC}.
% if advantage of DiffTune comparing to other differentiable-MPC based parameter learning-approaches is needed, then refer to the contribution statement of DiffTune-MPC paper.

%% file: approach.tex
\section{DiffTune setup for locomotion MPC}
In this section, we apply DiffTune to learn the optimal parameters of HECTOR's locomotion MPC controller under the scenarios of walking with different trajectories. We will introduce the choice of loss function and differentiable dynamics in the sequel.

\subsection{Loss function design}
% In nominal bipedal robots' MPC parameters tuning, researchers and engineers always suffer from the trade-off between tracking performance and control smoothness. Gaining better tracking performance always requires more violent control inputs, which can lead to greater energy consumption and serious jittering motions. To address this issue, we consider both tracking performance and control smoothness in designing our loss function. Our loss function can be written as:

We use the following loss function to guide the parameter optimization by DiffTune:
\begin{equation} \label{eq: loss function}
      L = \alpha_1  L_{\text{Eul}} + \alpha_2  L_{\text{pos}} +  L_{\text{ctrl}},
\end{equation}
where $L_{\text{Eul}}= \sum_{j=1}^{T} \norm{ \mathbf{e}_j -\mathbf{e}^{\text{ref}}_j}^2$ and $L_{\text{pos}} = \sum_{j=1}^{T} \norm{ \bm{p}_{c,j} - \bm{p}^{\text{ref}}_{c,j}}^2$ accounts for the tracking error in the float-base attitude and position, respectively; $L_{\text{ctrl}} = \sum_{j=1}^{T} \norm{ \bm{u}_j - \bm{u}_{j-1}}^2$ accounts for the smoothness of consecutive control actions. The weights $\alpha_1$ and $\alpha_2$ are hyperparameters to tradeoff between tracking performance and control smoothness (acquiring better tracking performance tends to require more violent control inputs, which can lead to serious jittering), as well as balancing the losses of different units and orders of magnitude.

\subsection{Choice of dynamics for differentiable programming}

A seemingly ideal choice of 
dynamics to be applied for parameter learning (i.e., $f(\cdot)$ in problem~\eqref{prob: controller tuning as a parameter optimization}) via differentiable programming is the full-order dynamics~\eqref{eq:fullDynamics}, since it is a high-fidelity
description of the robot's full-body motion with respect to the actuations. 
However, it is not a suitable choice with differentiable-programming-based learning for the following two reasons: (i) The full-order dynamics~\eqref{eq:fullDynamics} models a floating-base rigid-body system with actuation purely in the joint space (i.e., joint torques). When modeling a legged robot, this choice of control interface additionally requires the ground contact model to simulate interactions (i.e., constraint forces) at contact points, which introduces discrete variables into the dynamics and hence violates the assumption of differentiable dynamics by differentiable programming (and DiffTune).
(ii) The goal of this work is to learn the optimal control parameters for the MPC stance controller.
If one applies the full-order dynamics~\eqref{eq:fullDynamics}, then the swing leg's motion and joint control are also needed for propagating such dynamics. The swing leg requires a separate inverse kinematics solution to map foot positions into joint space commands, where the solution is often obtained through iterative programming, whose differentiation requires unrolling~\cite{scieur2022curse}. Hence, this procedure complicates the computation. 
These limitations indicate that the full-order dynamics~\eqref{eq:fullDynamics} is not a suitable choice for parameter learning for a stance controller with differentiable programming.

Following the reasons listed above, the SRBM~\eqref{eq:srbd_eom} is well-suited for differentiable learning: the dynamics itself is differentiable, and the control interface (using GRFM) matches the locomotion controller's output.
However, since the SRBM itself is simplified (with underlying assumptions on the negligible weights of lower thigh and calf links and constant MoI of the upper body and hips), performing iterative parameter learning with this dynamical model in DiffTune will introduce distribution shifts~\cite{cheng2022difftune}, which reduces the applicability of the learned parameters as the number of iterations grows.

To tackle these issues, we introduce a Ground Reaction Force-and-moment Network (GRFM-Net) to enhance the fidelity of the bipedal SRBM in a data-driven manner, thus allowing for the augmented model to closely approximate the physical system. % with full robot dynamics, contact dynamics, and associated low-level controllers while preserving the differentiability for the ease of parameter learning.
For the sake of generality, we use the following control-affine system to represent the commonly used dynamical model in simulation:
\begin{equation}\label{eq: nominal dynamics}
    \boldsymbol{x}_{j+1} = f(\boldsymbol{x}_j) + g(\boldsymbol{x}_j) \boldsymbol{u}_j, \ \boldsymbol{x}_0 \text{ given}.
\end{equation}
The nominal dynamics \eqref{eq: nominal dynamics} govern the major evolution of the system, where $\boldsymbol{u}_j$ denotes the control action produced by a controller. However, the nominal model \eqref{eq: nominal dynamics} cannot fully capture the evolution of the states on a physical robot due to factors that are hard to model or parameters that may be challenging to identify, which constitutes a major cause for the sim-to-real gap. Among them, one of the major sources of the sim-to-real gap is caused by the discrepancy between commanded control action $\boldsymbol{u}$ (as the output of the controller) and control effect $\bar{\boldsymbol{u}}$ (as the actual force/moment/actuation that drives the system). Specifically, in the context of legged robots the control effect $\bar{\boldsymbol{u}}$ is altered from the commanded control action $\boldsymbol{u}$ due to the physics of the low-level actuators and motors, as well as complex contact physics with the ground. We treat the control effect $\bar{\boldsymbol{u}}$ as a mapping $\phi(\cdot)$ of the control action and its history, i.e., $\bar{\boldsymbol{u}}_j = \phi(\boldsymbol{u}_{j-w:j})$, where $w$ is the length of historic control actions that influence the current control effect. This  results in the augmented dynamics that describe the evolution of the actual system state $\bar{\boldsymbol{x}}$, i.e., 
\begin{equation}\label{eq: physical system dynamics}
    \bar{\boldsymbol{x}}_{j+1} = f(\bar{\boldsymbol{x}}_j) + g(\bar{\boldsymbol{x}}_j) \phi(\boldsymbol{u}_{j-w:j}), \ \bar{\boldsymbol{x}}_0 \text{ given}.
\end{equation}

Due to the heavily non-linear and time-varying properties of the effects we mentioned above, modeling how the control command is turned into the real control effect is not an easy task. 
% \textcolor{red}{In our case, the GRFM generated by locomotion MPC has a strong assumption of simplified modeling. However, this optimized GRFM compared with the actual GRFM observed on the real robot or high-fidelity simulator has noticeable variations especially in dynamic motions, further limiting the performance of SRBM-based control strategies.}
In terms of the HECTOR robot, when executing the MPC-generated GRFM (corresponding to $\boldsymbol{u}$), due to the unavoidable mechanical transmission backlash and frictions, the actual GRFM exerted from the ground to the robot (corresponding to $\bar{\boldsymbol{u}}$) has noticeable variations compared to the MPC-commanded GRFM control action. 
Therefore, we propose to learn the mapping $\phi$ in a data-driven approach, where a deep neural network (DNN) $\hat{\phi}$ can approximate the mapping $\phi$, and we name the network by GRFM-Net.
More importantly, the DNN can be put back into the simulator (for differentiable learning), which will elevate the fidelity of the simulation by augmenting the actuation model. The augmentation yields a simulated system that behaves more accurately like the physical system given the same control commands (than the original simulated system). Moreover, existing key differentiable features of the nominal dynamics \eqref{eq: nominal dynamics} can be preserved and leveraged for DiffTune. 
Our idea of GRFM-Net is from the actuator network (AN) \cite{hwangbo2019learning}. However, unlike AN, which only considers a short chain of discrepancy between motor command (position and speed) and motor output (torque), our mapping from the MPC-produced GRFM to the actual GRFM contains a longer chain, including contact interaction, mechanical system behaviors, and the low-level actuators' own dynamics.

\subsection{Data collection and training of the GRFM-Net}
Due to the limited sensor precision and hardware design challenges within constrained spaces, most legged robots do not have accurate force sensors at the foot. To mitigate the issue in data collection, we built a MATLAB/Simulink-based high-fidelity simulator (HFS) equipped with accurate ground effect sensors at the ground contact points. In HFS, the bipedal robot dynamics model was built with the Simscape Multibody library and was represented as a multi-link rigid-body system with elastic ground contact. It is capable of simulating mechanical properties such as backlash, joint damping, and linkage friction to offer more realistic mechanical behavior. We collect the data pairs of control commands and control effects (both 12-dimensional vectors, including 3-dimensional forces and moments at both feet) from HFS under various reference trajectories, making the GRFM measurements a good representative of the real ground reaction (and the label of GRFM-Net training).  

We separate force and moment data as they have different physical units and train two GRFM-Nets in the same training setting. Each GRFM-Net is a multi-layer perception with 18 inputs (history length $w=3$), three hidden layers (each with 64 neurons), and six outputs. We use the softsign activation function~\cite{hwangbo2019learning} and conduct layer normalization. We use Adam optimizer and the learning rate is set to 0.001. To avoid overfitting, we use an L2 regularization with a weight of 0.001. The two GRFM-Nets reach the plateau after training 200 epochs, both in training and validation losses. The final MSE loss for force and moment GRFM-Nets are 12.78 and 0.55, respectively.

%% file: results.tex
\section{Experimental Results}
\subsection{Fidelity check on the Augmented Actuator Network}
\begin{figure}[!h]
\vspace{0.2cm}
\centering
\includegraphics[clip, trim=0cm 2.3cm 3.3cm 0cm, width=1\columnwidth]{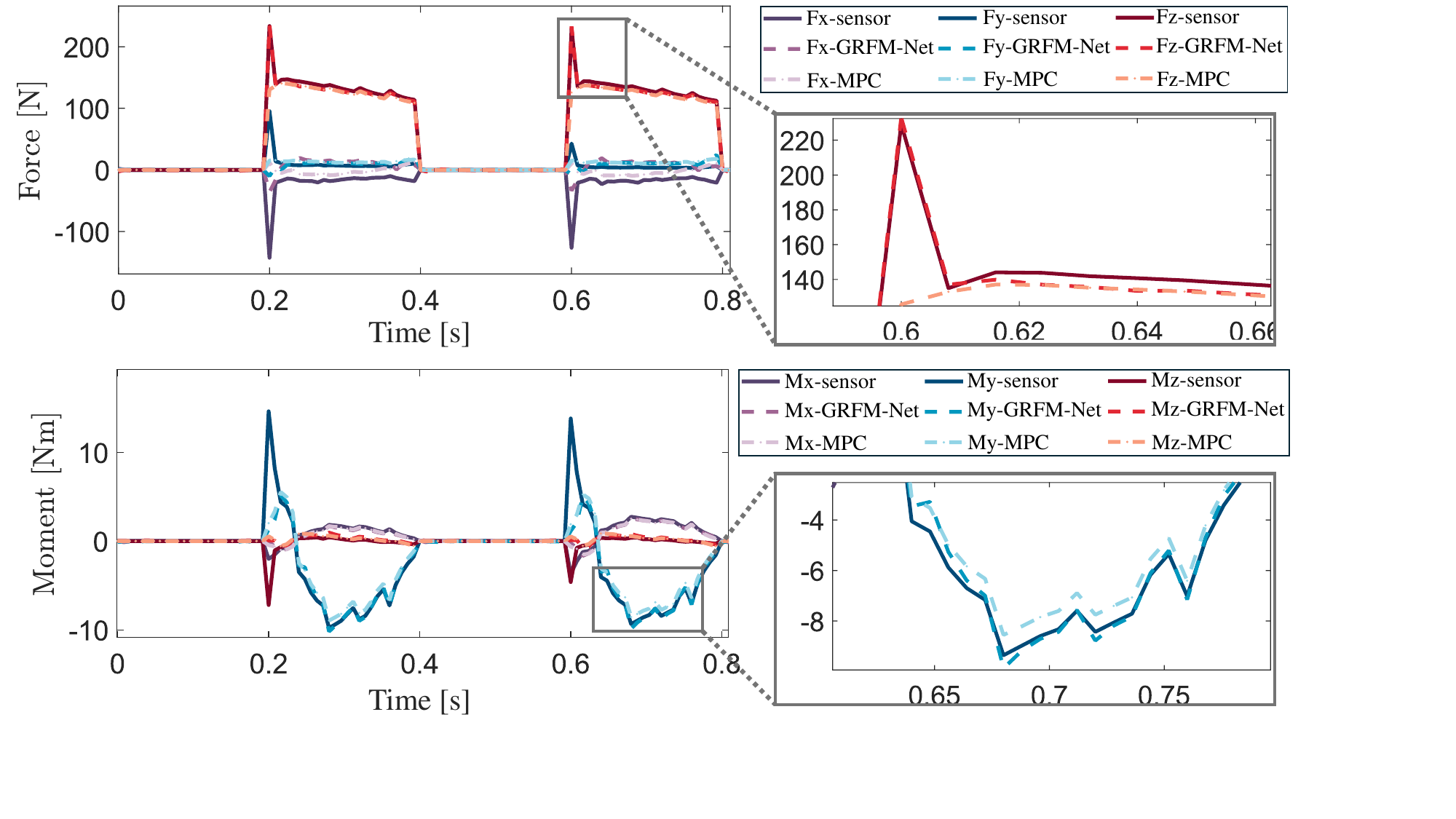}
\caption{Force and moment profiles for GRFM-Net fidelity check}
\label{fig: AAN fidelity}
\vspace{-0.4cm}
\end{figure}

We first examine the improvement in fidelity by the GRFM-Net. 
We test the straight-line walking in the HFS, recording force and moment of (i) MPC output, (ii) GRFM-Net output, and (iii) HFS sensor readings. Figure~\ref{fig: AAN fidelity} shows the left foot force and moment profile in the first two steps. 
It can be observed that the gap between sensor readings and GRFM-Net output is smaller than the gap between sensor readings and MPC output, indicating that GRFM-Net is effective in capturing the discrepancies between the control commands and control effects and hence improving simulation fidelity. % and, hence, mitigating the sim-to-real gap. 

\subsection{Hardware test results} 
We select three trajectories --- straight line, C-shape curve, and S-shape curve --- with increasing levels of control challenge for testing the effectiveness of GRFM-Net-involved DiffTune. 
For each trajectory, we test HECTOR's walking performance under four sets of MPC parameters, including those found by DiffTune with GRFM-Net (named by ``the proposed'') and three baselines: % (diagonal entries of matrices Q, R in \ref{eq:MPCcost}). 
(i) nominal untuned MPC parameters (for showing the tracking performance improvement by the proposed compared with initial parameters), (ii) expert-tuned MPC parameters (for comparing the proposed parameters with best manually tuned parameters), and (iii) parameters by DiffTune without GRFM-Net (as ablation to examine the role of GRFM-Net in addressing the sim-to-real gap). 
% \textcolor{blue}{will massage this part: We compared proposed method with (i) to show that DiffTune can improve bipedal robot's controller tracking performance without scarifying too much control smoothness. Moreover, compared with (ii), we demonstrated that DiffTune can tune better MPC parameters than human expert in terms of bipedal robot total walking loss. (iii) works as an ablation test to address the importance of AAN. Without AAN, the severe sim-to-real gap would largely impair the ability of DiffTune, making DiffTune not as effective as designed.} 
For each trajectory, we ran DiffTune with and without GRFM-Net for ten iterations, with parameters initiated from the nominal untuned MPC parameters. 
% The differentiation of the MPC locomotion follows~\cite{tao2023DT-MPC}.
We used the following hyperparameters in all cases: the weights\footnote{We choose the values of $\alpha_1$ and $\alpha_2$ such that the order of magnitude of the three gradients $\partial L_{\text{Eul}} / \partial \boldsymbol{\theta}$, $\partial L_{\text{pos}} / \partial \boldsymbol{\theta}$, and $\partial L_{\text{ctrl}} / \partial \boldsymbol{\theta}$ are aligned.} in the loss function are $\alpha_1 = 1\times 10^5$ and $\alpha_2 = 2\times 10^5$; the learning rates are $\beta_Q = 0.05$ and $\beta_R = 0.08$ for $Q$ and $R$ matrices in the MPC cost function.  
At test time, we deployed the four sets of MPC parameters on the HECTOR V2 bipedal robot to walk each trajectory and recorded the robot's position, pose, and control actions at 1000 Hz. % In the following sections, we will demonstrate the hardware test results for each trajectory.

The three trajectories are defined as follows:

\noindent \textbf{Trajectory 1: Straight line}
\begin{equation}
    \begin{aligned}
         v_x^{\text{des}}  = 0.5 \ \mathrm{m}/\mathrm{s}, \quad
        \dot{\gamma}^{\text{des}}  = 0.
    \end{aligned}
\end{equation}

\noindent \textbf{Trajectory 2: C-shape curve}
\begin{equation}
        v_x^{\text{des}}  = 0.25 \ \mathrm{m}/\mathrm{s}, \quad
        \dot{\gamma}^{\text{des}} = 0.25\pi \ \mathrm{rad} / \mathrm{s}
\end{equation}

\begin{figure*}[!t]
\centering
\includegraphics[clip, trim=1.5cm 0.9cm 1.2cm 0.2cm, width=2.05\columnwidth]{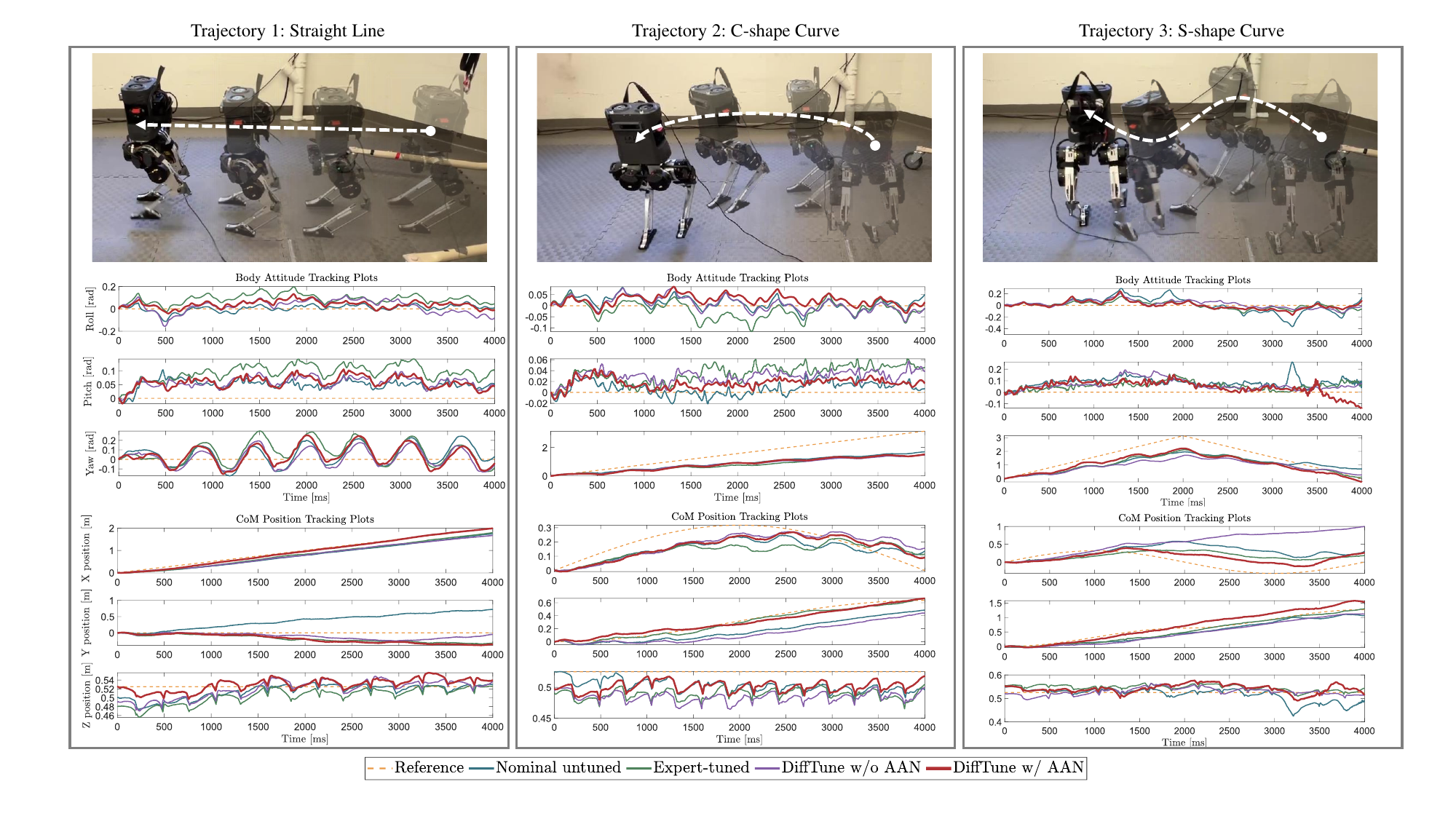}
\caption{Experimental snapshots of three trajectories and associated CoM position and Euler angles tracking plots.}
\label{fig: collective plots}
\vspace{-0.3cm}
\end{figure*}

\noindent \textbf{Trajectory 3: S-shape curve}
\begin{equation}
    \begin{aligned}
        v_x^{\text{des}} & = 0.5 \ \mathrm{m}/\mathrm{s} \\
        \dot{\gamma}^{\text{des}} & =\left\{\begin{array}{c} 0.5\pi \ \mathrm{rad} / \mathrm{s}, \ 0 \leq t \leq 2 \mathrm{~s} \\ -0.5\pi \ \mathrm{rad} / \mathrm{s}, \ 2<t \leq 4 \mathrm{~s} \end{array}\right.
    \end{aligned}
\end{equation}
The illustration of the three trajectories is shown in Fig.~\ref{fig: collective plots}, as well as the plots for pose and position tracking by the different MPC parameters.
The losses are shown in Table~\ref{table: all traj loss}.

DiffTune with GRFM-Net found MPC parameters that reduce the weighted total loss $L$ by 68.7\%, 29.7\%, 41.9\% compared with the nominal untuned parameters and 19.4\%, 40.5\%, 17.0\% compared with the expert-tuned parameters, demonstrating the optimality of the learned parameters. %\sout{In terms of control loss $L_{\text{ctrl}}$ that measures the smoothness of the control actions, DiffTune with AAN achieves 20\%, 40.5\%, and 17\% reduction compared to the expert-tuned parameter and yields visually smoother motions in experiments.} 
In terms of position and attitude tracking, the parameters learned by DiffTune and GRFM-Net achieve better accuracy than the expert-tuned parameters, with a marginally higher control loss $L_{\text{ctrl}}$ (with increase no more than 9\%) than the latter. 
Note that the parameters by DiffTune without GRFM-Net resulted in a large total loss, especially when running challenging trajectories (C- and S-shape curves). 
This result confirms that the sim-to-real gap can be a serious issue, leading to unsatisfactory performance on the physical system, whereas the GRFM-Net can successfully mitigate the gap by augmenting the simulator to closely approximate the physical system.

\renewcommand{\arraystretch}{1} % Default value: 1
  \captionsetup{%size=footnotesize,
	%justification=centering, %% not needed
	skip=5pt, position = bottom}
\begin{table*}% [h!]
\vspace{0.2cm}
\caption{Loss evaluation in experiments with different parameters over the three testing trajectories. The loss functions are defined in \eqref{eq: loss function}.}
\centering
\setlength{\tabcolsep}{3pt}
\begin{small}
\begin{tabular}{ccccccccccccc}
\toprule 
{Trajectory} & \multicolumn{4}{c}{Straight line} & \multicolumn{4}{c}{C-shape curve}  & \multicolumn{4}{c}{S-shape curve}                                                                                     \\ \cmidrule(r){1-1} \cmidrule(r){2-5} \cmidrule(lr){6-9} \cmidrule(l){10-13}
                     \multirow{2}{*}{Loss}   & $L_{\text{Eul}}$ & $L_{\text{pos}}$  & $L_{\text{ctrl}}$ & $L$ & $L_{\text{Eul}}$ & $L_{\text{pos}}$  & $L_{\text{ctrl}}$ & $L$ & $L_{\text{Eul}}$ & $L_{\text{pos}}$  & $L_{\text{ctrl}}$ & $L$ \\ 
                     & ($\times 10^2$) & ($\times 10^2$) &  ($\times 10^5$) & ($\times 10^8$)  & ($\times 10^3$) & ($\times 10^2$) &  ($\times 10^5$) & ($\times 10^8$)  & ($\times 10^3$) & ($\times 10^2$) &  ($\times 10^6$) & ($\times 10^8$)\\
                     \midrule
Nominal Untuned      & 9.66  & 9.01  & \textbf{7.15}  & 2.78 
                     & 2.32  & 1.54  & \textbf{3.56}  & 2.63   
                     & 1.91  & 8.05  & 0.89  & 3.53   \\
Expert-tuned         & 2.49  & 4.09  & 8.11  & 1.08   
                     & 2.97  & 0.66  & 4.49  & 3.11   
                     & 1.65  & 4.06  & 1.00  & 2.47   \\
DiffTune w/o GRFM-Net     & \textbf{1.78} & 3.87  & 7.77  & 0.96  
                     & 3.01  & 2.38  & 4.76  & 3.50 
                     & 2.74  & 23.27 & \textbf{0.73}  & 7.41  \\
DiffTune w/ GRFM-Net      & 1.86  & \textbf{3.38} & 8.90  & \textbf{0.87}  
                     & \textbf{1.75} & \textbf{0.47} & 4.56 & \textbf{1.85} 
                     & \textbf{1.47}  & \textbf{2.86} & 1.02 & \textbf{2.05}   \\  
\bottomrule
\end{tabular}
\end{small}
\label{table: all traj loss}
\end{table*}

%% file: conclusion.tex
\section{Conclusions}
In this work, we applied DiffTune to the HECTOR robot to optimize locomotion MPC controller parameters for bipedal walking across multiple trajectories. We addressed the low-fidelity limitations of the differentiable SRBM by introducing the GRFM-Net, which enhances model fidelity by mapping MPC control actions to actual control effects. Trained using data from a high-fidelity simulator, the GRFM-Net's fidelity improvement was verified when integrated with the SRBM. Experiments with the HECTOR robot on various trajectories show that DiffTune, combined with GRFM-Net, improves tracking accuracy and maintains smooth control, reducing total loss by up to 40.5\% compared to expert-tuned MPC parameters. These results highlight the critical role of GRFM-Net in closing the sim-to-real gap for differentiable-programming-based parameter learning, especially in challenging locomotion tasks. Furthermore, the proposed data collection and training framework is generalizable to other platforms, rendering more efficient learning while holding sufficient fidelity.
% \textcolor{green}{Future work will look into ...}
% \jli{[some (maybe intangible) ideas]Future work will explore the possibility of expanding Difftune to (1) conduct parameter-tuning in an online fashion or (2) exploit its usage in full order and contact dynamics in the context of task-tailored MPC or trajectory optimization...}
% \textcolor{blue}{Qianzhong: possible future work: (1) Dive deeper into formal analysis of the trade-off between control effectiveness and control smoothness. Moreover, find a solution that can increase the tracking performance without sacrificing control smoothness with DiffTune. (2) Build a pipeline that tune the MPC parameters with DiffTune under multiple trajectories, mitigating the parameter's overfit to each trajectory when tuned only on one.}

Ongoing work involves improving the robustness of DiffTuned parameters under disturbances during deployment. Our future work will focus on expanding DiffTune to enable online parameter tuning and applying it to full-order and contact dynamics for more generalized usage. Additionally, we look to develop a pipeline for auto-tuning MPC parameters across multiple trajectories to improve generalization.